\documentclass[runningheads]{llncs}
\usepackage[T1]{fontenc}
\usepackage{algorithm}
\usepackage{algpseudocode}
\usepackage{amsmath}
\usepackage{graphicx,verbatim}
\usepackage{booktabs}
\usepackage{float}
\usepackage{enumitem}
\usepackage{multirow}

\usepackage{marvosym} 
\usepackage[colorlinks]{hyperref}

\begin{document}
\title{QCAgent: An agentic framework for quality-controllable pathology report generation from whole
slide image}
\titlerunning{QCAgent for quality-controllable WSI report generation}
%

\author{
Rundong Wang\inst{1,2}$^\dagger$ \and
Wei Ba\inst{3}$^\dagger$ \and
Ying Zhou\inst{1,2} \and
Yingtai Li\inst{1,2} \and
Bowen Liu\inst{1,2} \and
Baizhi Wang\inst{1,2} \and
Yuhao Wang\inst{1,2} \and
Zhidong Yang\inst{4} \and
Kun Zhang\inst{1,2} \and 
Rui Yan\inst{1,2} $^{\href{mailto:yanrui@ustc.edu.cn}{\textrm{\Letter}}}$ 
\and Shaohua Kevin Zhou\inst{1,2}$^{\href{mailto:skevinzhou@ustc.edu.cn}{\textrm{\Letter}}}$ 
}

\authorrunning{R. Wang et al.}

\institute{
School of Biomedical Engineering, Division of Life Sciences and Medicine, University of Science and Technology of China, Hefei, Anhui, 230026, P.R. China 
\and
Center for Medical Imaging, Robotics, Analytic Computing \& Learning (MIRACLE), Suzhou Institute for Advance Research, USTC, 215123, P.R. China
\and
Chinese PLA General Hospital Ninth Medical Center, China
\and
Hong Kong University of Science and Technology (HKUST), Hong Kong, China
}

\maketitle              
\footnote{\textsuperscript{$\dagger$}Equal contribution. \textsuperscript{\Letter}Corresponding authors.}

\begin{abstract}

Recent methods for pathology report generation from whole-slide image (WSI) are capable of producing slide-level diagnostic descriptions but fail to ground fine-grained statements in localized visual evidence. 
Furthermore, they lack control over which diagnostic details to include and how to verify them. 
Inspired by emerging agentic analysis paradigms and the diagnostic workflow of pathologists,   
who selectively examine multiple fields of view, we propose \textbf{QCAgent}, an agentic framework for quality-controllable WSI report generation. The core innovations of this framework are as follows: (i) it incorporates a customized critique mechanism guided by a user-defined checklist specifying required diagnostic details and constraints; (ii) it re-identifies informative regions in the WSI based on the critique feedback and text-patch semantic retrieval, a process that iteratively enriches and reconciles the report. Experiments demonstrate that by making report requirements explicitly prompt-defined, constraint-aware, and verifiable through evidence-grounded refinement, QCAgent enables controllable generation of clinically meaningful and high-coverage pathology reports from WSI.


\end{abstract}

\section{Introduction}
\label{sec:introduction}

Pathology foundation models  \cite{GPFM,TANGLE,PANTHER,CHIEF,GigaPath} have significantly advanced whole slide image (WSI) analysis and its applications. Among its applications, WSI-to-Text generation promises to reduce the pathologist's burden and enhance diagnostic consistency by generating descriptions \cite{PRISM,WsiCaption,HistGen,pathchat}. Unlike general image captioning, however, a clinical standard report must not only provide a final diagnosis but also rigorously cover specimen context, microscopic morphological descriptors (e.g., nuclear pleomorphism), and critical predictors such as lymphovascular invasion (LVI).

Due to the gigapixel scale of WSIs, current generative models (e.g., HistGen \cite{HistGen}, SGMT \cite{SGMT} and PRISM \cite{PRISM}) predominantly follow a single-pass paradigm, leading to two intractable issues. First, feature collapse occurs: to maintain computational feasibility, the model compresses thousands of image patches into compact global vectors. This drastic dimensionality reduction, while capturing general lesions, inevitably erases spatially sparse but diagnostically decisive features, such as micro-lesions or evidence of LVI \cite{WSI-VQA}. Second, linguistic prior bias arises: lacking a ``visual look-back'' mechanism, Large Language Models (LLMs) tend to hallucinate generic clinical commons (e.g., ``cellular atypia'') based on diagnostic labels (e.g., ``gastric cancer'') when global features are insufficient for detailed synthesis. Such details lack correlation with specific visual evidence from the current slide and do not address missing information required for clinical quality control, severely undermining the clinical credibility of reports \cite{SlideChat}.

To bridge the gap between global context and local evidence, we draw inspiration from how pathologists iteratively navigate between low-power screening and high-power verification. This workflow aligns with emerging agentic paradigms \cite{agentbench,sapkota2025ai,wang2024beyond,truhn2026artificial,zhao2026agentic}. Although recent work has explored agents in computational pathology \cite{cpathagent,ghezloo2025pathfinder,pathagent,GIANT,SurvAgent,WSI-Agents}, agentic \emph{on-demand back-tracking} for controllable, evidence-grounded report generation remains underexplored.

Motivated by this, we propose QCAgent, an agent-based, quality-controllable pathology report generation framework. First, the system employs a WSI foundation model, PRISM \cite{PRISM}, to extract global semantic features, supplemented by representative patch information via a cluster-based random sampling strategy. Subsequently, it invokes a pathology-expert reasoning model, Patho-R1 \cite{Patho-R1}, to generate an initial draft. Crucially, we introduce a quality control module that utilizes advanced LLMs \cite{LLMBenchmarking,Qwen,qwen3vl} to simulate a pathologist's auditing behavior. It identifies critical missing information, converts them into targeted retrieval instructions, and triggers Patho-R1 to revise the report. This process iterates until the report passes the QC criteria. This mechanism enables evidence-based report generation by explicitly aligning visual findings with clinical narratives. The main contributions of this paper are three-fold:


%
\begin{itemize}[label=\textbullet,leftmargin=*,noitemsep,topsep=0pt,parsep=0pt]
    \item \textbf{Iterative QC architecture.} We reformulate report generation as a multi-round \emph{audit--retrieve--revise} process. A structured QCAgent automatically identifies non-compliant statements and missing clinical information, enabling closed-loop refinement.
    \item \textbf{Demand-driven evidence retrieval.} We introduce a \emph{text-guided evidence retrieval} mechanism. Instead of random sampling, the QCAgent converts missing fields into morphology-oriented queries to retrieve the most relevant visual patches from the WSI, grounding the report completion in targeted evidence.
    
    \item \textbf{Strict anti-hallucination design.} We adopt an evidence-priority scheme (supplemental visual evidence $>$ initial draft) to reduce hallucinations.
    By explicitly separating verifiable from non-verifiable fields, only evidence-supported claims are retained in the final report.
\end{itemize}

\section{Method}

\begin{figure}[tbp]
    \centering
    \includegraphics[width=\textwidth]{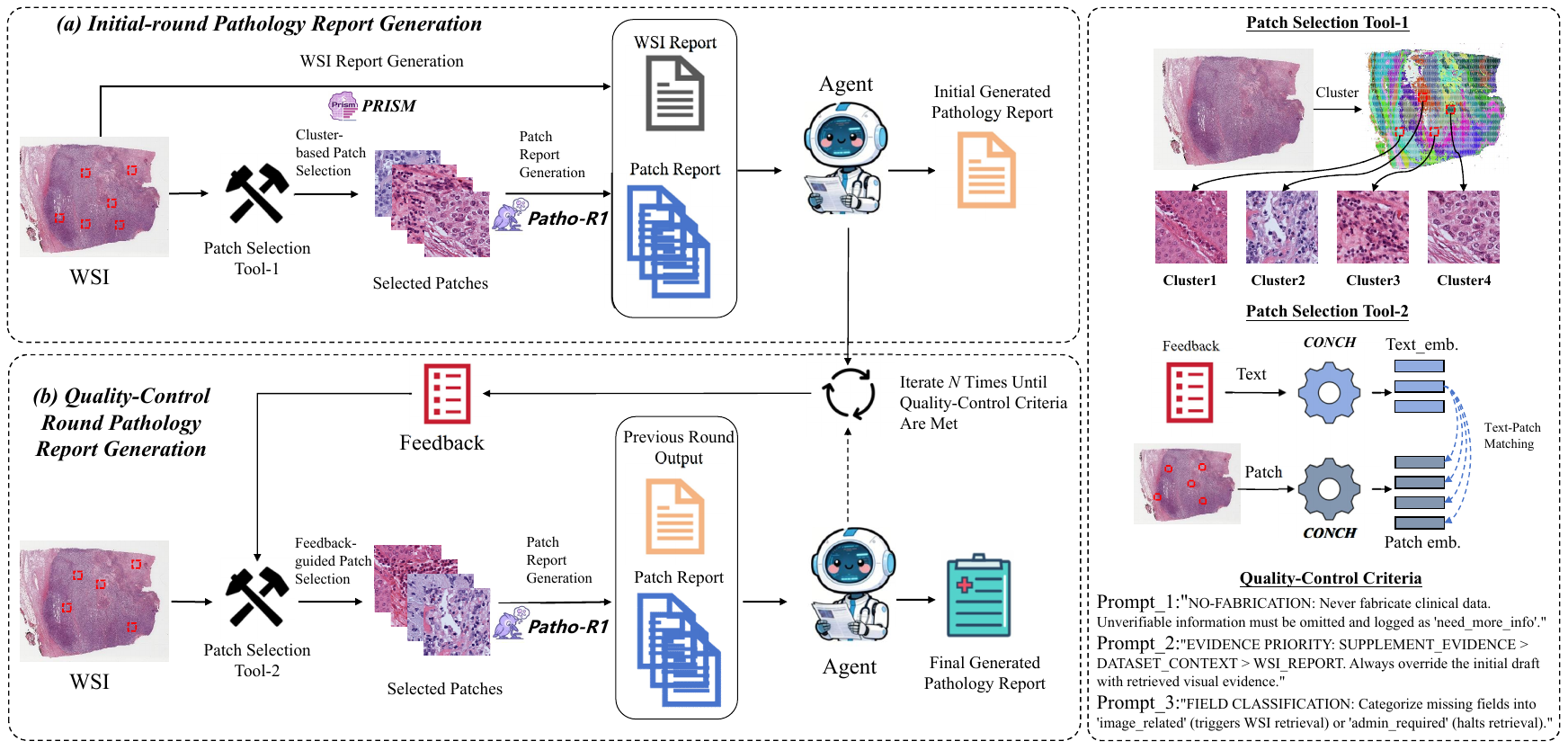}
    \caption{Pipeline of QCAgent: (a) initial-round report generation using PRISM and cluster-based patch selection (Tool-1) with Patho-R1 patch reports; (b) QC-round iterative refinement using feedback-guided CONCH retrieval (Tool-2) and Patho-R1 evidence until QC criteria are met.}
    \label{fig:workflow}
\end{figure}

This section elaborates on the proposed QCAgent, an agent-based, quality-controllable framework for pathology report generation from Whole Slide Images (WSIs), as illustrated in Fig.~\ref{fig:workflow}. The overall computational pipeline of this framework consists of two core stages: (a) \textbf{Initial-round generation}, which aims to rapidly construct a WSI-level draft and extracts local visual evidence with global coverage; and (b) \textbf{Quality-Control (QC) rounds of iterative generation}, which translates information gaps within the report into explicit ``evidence demands''. Through a closed-loop mechanism of ``feedback-guided evidence retrieval'' and ``evidence-driven revision'', this stage progressively fulfills stringent quality control criteria.

\paragraph{Problem Formulation.}
Given a whole-slide image (WSI) $S$ and a set of pre-extracted patch features $\mathcal{F}=\{f_1,\ldots,f_N\}$ obtained by CONCH\cite{CONCH}, together with an initial draft report $r_0$ produced by a foundation model, our goal is to generate an informative and accurate pathology report $r^*$ through $T$ rounds of iterative refinement. We formulate the process as a multi-round optimization with a maximum of $T$ iterations:

\begin{equation}
r^* = \mathrm{QC\text{-}Iterate}(r_0, \mathcal{F}, S; T),
\end{equation}
where each iteration comprises four consecutive stages: (i) quality-control (QC) assessment, (ii) evidence retrieval, (iii) visual diagnosis, and (iv) report revision.

We define \emph{quality-controllability} as the capability of the system to explicitly follow and enforce a predefined set of quality control criteria $\mathcal{K}$ during inference. These criteria include, but are not limited to, a report schema, required fields, anti-hallucination rules, evidence-priority principles, and standardized writing templates. At each iteration, the system outputs structured QC results, enabling users to dynamically configure $(\mathcal{K},T)$ to achieve a controllable trade-off between generation quality and computational cost. Under this formulation, ``controllability'' is not merely post-hoc scoring; instead, hard constraints are embedded into an iterative ``audit--evidence seeking--revision'' closed loop, yielding a transparent and auditable reasoning trace.

Concretely, our QC prompts implement three core rules:
\begin{description}
    \item \textbf{NO-FABRICATION:} Never fabricate clinical data. Any unverifiable information must be omitted from the report and explicitly logged as \texttt{need\_more\_info}.
    \item \textbf{EVIDENCE PRIORITY:} When supplemental evidence conflicts with the initial draft, revisions follow the priority order $\text{SUPPLEMENT\_EVIDENCE} > \text{DATASET\_CONTEXT} > \text{WSI\_REPORT}$.
    \item \textbf{FIELD CLASSIFICATION:} Missing fields are categorized into \texttt{image\_related} and \texttt{admin\_required}.
\end{description}

\paragraph{Initial Round: Global Draft and Coverage-Oriented Evidence Extraction.}
As illustrated in Fig.~\ref{fig:workflow}(a), the initial round consists of two parallel streams:
\begin{enumerate}[leftmargin=*,noitemsep,topsep=0pt,parsep=0pt]
    \item \textbf{WSI-level report generation (PRISM).} We first employ a pretrained WSI-level foundation model to produce an initial draft report $r_0$, which serves as a global narrative backbone for subsequent refinement.
    \item \textbf{Cluster-based patch selection (Patch Selection Tool-1).} To provide representative \emph{coverage-oriented} evidence in the initial round, we perform clustering in the feature embedding space over candidate patches and uniformly sample from each cluster to construct a representative patch set $\mathcal{P}^{(1)}$. This strategy ensures comprehensive initial information selection.
    \item \textbf{Patch-level report generation (Patho-R1).} The selected patches $\mathcal{P}^{(1)}$ are then fed into the pathology vision-language model (VLM) Patho-R1 to generate patch-level descriptions $\mathcal{E}^{(1)}$, which serve as detailed evidence. In practice, Patho-R1 is executed in batches and produces fine-grained diagnostic cues, providing direct visual evidence to complement the WSI-level draft.
\end{enumerate}

Finally, the agent fuses the global draft and local evidence to produce the initial generated report $r_0'$, which is the starting point for the subsequent QC iterations.

\paragraph{Quality-Control Rounds: Feedback-Guided Evidence Seeking and Evidence-Driven Revision.}
As shown in Fig.~\ref{fig:workflow}(b), the QC rounds take the \emph{previous-round report} together with accumulated evidence and the quality criteria as inputs, and perform closed-loop refinement with a maximum of $T$ iterations.

At iteration $t$, the QCAgent conducts a fine-grained, structured audit of the previous report $r_{t-1}$ and outputs: (i) a set of missing key fields $\mathcal{M}_t$, (ii) a set of morphology-driven retrieval queries $\mathcal{Q}_t$, and (iii) a revised draft $r_t'$. The query set $\mathcal{Q}_t$ follows a demand-driven principle (``retrieve what is missing''). For example, when lymphovascular invasion (LVI) is missing, the agent generates a morphology-oriented query such as \emph{``Histopathological evidence for lymphovascular invasion in gastric adenocarcinoma.''} Meanwhile, the QCAgent enforces the \textbf{NO-FABRICATION} rule: for any field that cannot be supported by reliable visual evidence from the WSI, the system refuses to hallucinate and explicitly logs it as \texttt{need\_more\_info}.

Tool-2 performs text-guided retrieval in the CONCH's embedding space:
\begin{enumerate}
    \item \textbf{Text encoding:} Compute the normalized query embedding $q=\mathrm{CONCH}_{\text{text}}(\mathcal{Q}_t)$;
    \item \textbf{Similarity search:} Retrieve TopK patches via cosine similarity using a FAISS IndexFlatIP index, i.e., $\mathrm{TopK}(\{q\cdot f_i\}_{i=1}^{N})$;
    \item \textbf{De-duplication:} Exclude patches already retrieved in previous rounds to reduce redundancy;
    \item \textbf{Patch cropping:} Return the Top-$K$ patch coordinates and similarity scores, and crop the corresponding high-resolution patches from the original WSI.
\end{enumerate}

Compared with random sampling or global clustering alone, this demand-driven retrieval paradigm pinpoints tissue regions most relevant to the current QC feedback, substantially improving the efficiency of evidence acquisition.

The retrieved patches are then re-input to Patho-R1 to generate fine-grained descriptions, yielding a new set of supplemental evidence $\mathcal{E}_t$. This evidence is fed back into the subsequent QC loop as key supporting material for revision.
The agent produces the revised report $r_t$ by integrating $(r_{t-1}, \mathcal{E}_t, \mathcal{K})$. The iterative process terminates when any of the following conditions holds: (i) the report satisfies all QC criteria; (ii) the remaining missing fields are all non-image-evidenceable; or (iii) the iteration budget reaches $T$. 

\begin{algorithm}[tbp]
\caption{  Agent-Based Framework for Quality-Controllable}
\label{alg:qcagent}
\textbf{Input:} WSI $\mathbf{S}$, patch features $\mathcal{F}$, coordinates $\mathcal{C}$, quality criteria $\mathcal{K}$, max rounds $T$ \\
\textbf{Output:} Final generated report $r^*$
\begin{algorithmic}[1]
\State $r_0 \gets \mathrm{PRISM}(\mathbf{S})$
\State $\mathcal{P}^{(1)} \gets \mathrm{Tool\text{-}1\_ClusterSelect}(\mathbf{S})$
\State $\mathcal{E}^{(1)} \gets \mathrm{PathoR1}(\mathcal{P}^{(1)})$
\State $r_0' \gets \mathrm{AgentFuse}(r_0, \mathcal{E}^{(1)}; \mathcal{K})$
\For{$t = 1, \dots, T$}
    \State $(\mathcal{M}_t, \mathcal{Q}_t, r_t') \gets \mathrm{QC}(r_{t-1}, \mathcal{E}_{t-1}; \mathcal{K})$
    \If{$\mathrm{Pass}(\mathcal{M}_t, \mathcal{K})$}
        \State \textbf{break}
    \EndIf
    \State $\mathcal{P}^{(2)}_t \gets \mathrm{Tool\text{-}2\_CONCH\text{-}Retrieve}(\mathcal{Q}_t, \mathcal{F}, \mathcal{C})$
    \State $\mathcal{E}_t \gets \mathrm{PathoR1}(\mathcal{P}^{(2)}_t)$
    \State $r_t \gets \mathrm{Revise}(r_t', \mathcal{E}_t; \mathcal{K})$
\EndFor
\State \Return $r^* \gets r_t$
\end{algorithmic}
\end{algorithm}

\section{Experiments and Results}
\label{sec:experiments}

\subsection{Datasets and Evaluation Metrics}
\label{sec:datasets_metrics}
\paragraph{Datasets.} To assess generalization across clinical settings and languages, 
we evaluate our framework on two independent, large-scale cohorts of gastric cancer: 

\noindent \textbf{TCGA-STAD (public, English).}
This cohort contains high-resolution WSIs from stomach adenocarcinoma in The Cancer Genome Atlas (TCGA). For TCGA-STAD, we use the slide-level pathology reports from TITAN \cite{Titan} as ground-truth.

\noindent \textbf{In-house dataset (private, Chinese).}
This cohort consists of 524 H\&E WSIs. Ground-truth is the corresponding Chinese clinical pathology reports, which are highly structured and concise, emphasizing factual statements.

\paragraph{Evaluation metrics.}
We report both traditional text overlap metrics (BLEU-1/4, ROUGE-L, METEOR) and a semantic similarity metric (BERTScore). We also introduce a clinically motivated \emph{Field Recall}, which measures the proportion of clinically critical pathology fields successfully covered by the generated report. 

\label{sec:setup_baselines}
\paragraph{Baseline and implementation details.}
We adopt PRISM, a publicly available WSI-level foundation model, as the initial-report baseline. To validate the effectiveness of iterative quality control, we compare the one-pass draft generated by PRISM (without QC) against the final output of QCAgent.
Our setup employs the CONCH encoder for text retrieval (Tool-2), Patho-R1 for visual generation, and Qwen3-VL-30B-A3B-Thinking as the core QCAgent.  The QC loop runs for up to $T=3$ iterations. At each iteration, Tool-2 retrieves the top $K=3$ patches for each QC query, with previously retrieved patches excluded to reduce redundancy.


\subsection{Experimental Results}
\label{sec:main_results}

\textbf{Results on TCGA-STAD (English).}
\label{sec:main_results}
Compared with the PRISM baseline, our QCAgent achieves consistent and substantial improvements across all metrics on TCGA-STAD (Table~\ref{tab:main_results}). Notably, Field Recall increases from 32.08\% to 63.35\%, indicating that the iterative QC process successfully recovers clinically important missing findings. In addition, we observe a clear increase in report completeness: PRISM reports are typically brief, while QC-refined reports provide more detailed and comprehensive descriptions.



\begin{table}[t]
\centering
\caption{Quantitative results of WSI report generation on two datasets.}
\label{tab:main_results}
\resizebox{\textwidth}{!}{
\begin{tabular}{llrrrrrr}
\toprule
Dataset & Method & BLEU-1 & BLEU-4 & ROUGE-L & METEOR & FieldRecall & BERTScore \\
\midrule
\multirow{2}{*}{\parbox{2cm}{\raggedright TCGA-\\STAD}} 
& PRISM (Baseline) & 0.0986 & 0.0319 & 0.0947 & 0.0985 & 0.3208 & 0.1245 \\
& QCAgent (Ours) & 0.2857 & 0.0836 & 0.2784 & 0.4087 & 0.6335 & 0.2738 \\
\midrule
\multirow{2}{*}{In-house} 
& PRISM (Baseline) & 0.0226 & 0.0083 & 0.0484 & 0.0333 & 0.1908 & -0.0831 \\
& QCAgent (Ours) & 0.0924 & 0.0186 & 0.2490 & 0.2307 & 0.4863 & 0.2370 \\
\bottomrule
\end{tabular}
}
\end{table}

\noindent \textbf{Results on In-house Cohort (Chinese).}
\label{sec:main_results}
We further evaluate cross-lingual transfer by directly applying our QCAgent to the private Chinese cohort without any finetuning on Chinese pathology data. As shown in Table~\ref{tab:main_results}, the QCAgent yields marked improvements over PRISM across overlap-based metrics and semantic similarity. We note that the Chinese ground-truth reports are highly condensed, often in phrase-like formats, while our system generates more detailed narrative paragraphs. This stylistic mismatch can depress n-gram matching metrics. Nevertheless, the improvement in BERTScore-F1 suggests better alignment at the semantic and factual level.

\begin{figure}[t]
    \centering
    \includegraphics[width=\textwidth]{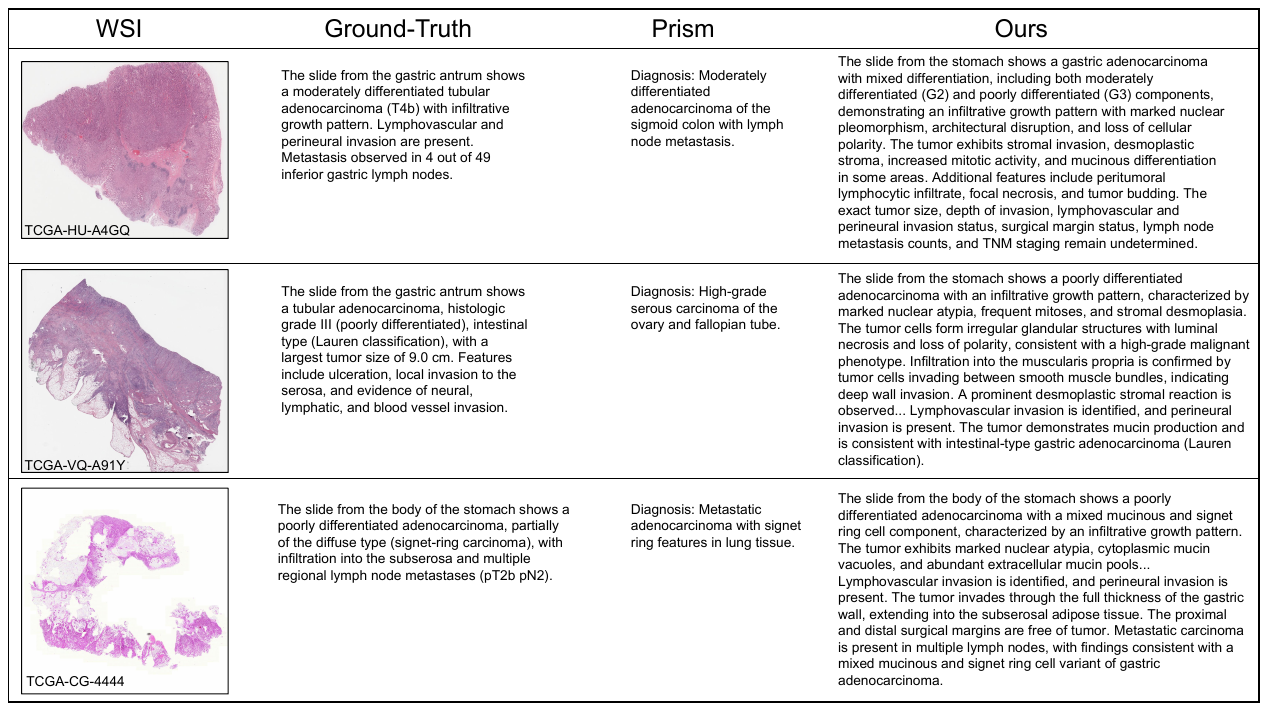}
    \caption{\textbf{Qualitative comparison of pathology reports on TCGA-STAD.} 
    For each case, we show the WSI thumbnail, the ground-truth clinical report, the one-pass draft generated by PRISM, and our QC-refined report.}
    \label{fig:qualitative}
\end{figure}

\begin{figure}[t]
    \centering
    \includegraphics[width=\textwidth]{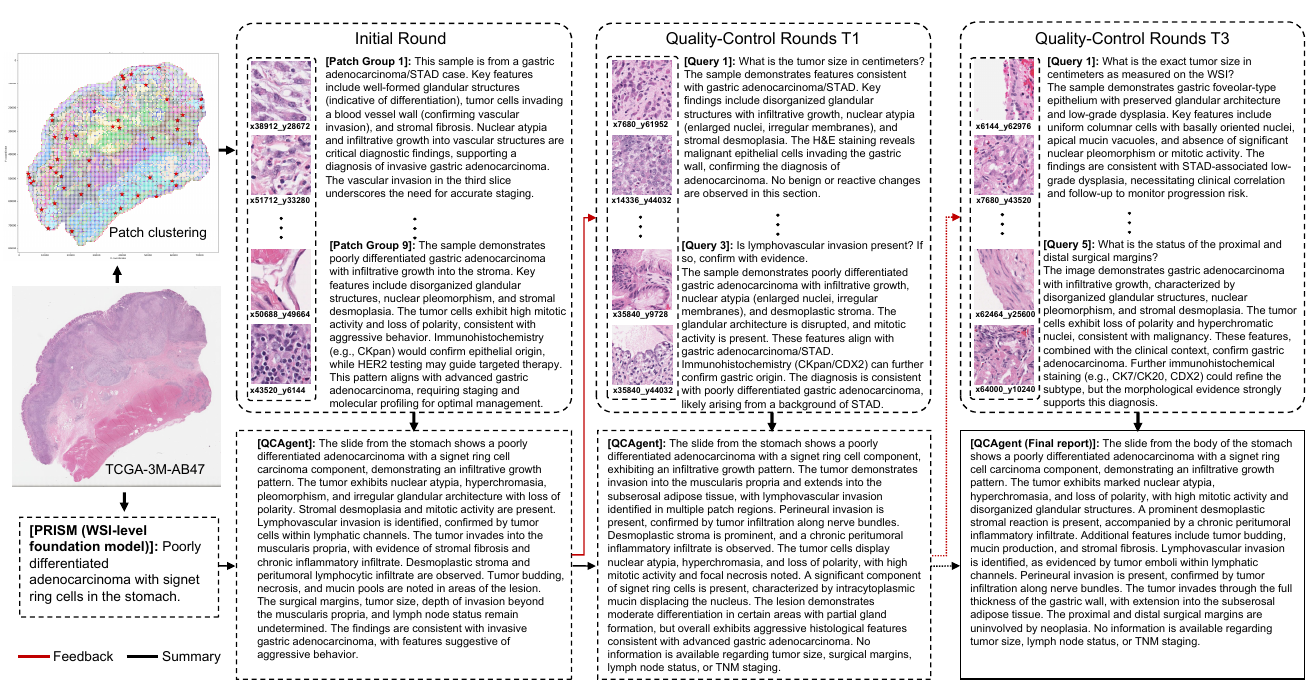}
    \caption{\textbf{Case study of the iterative audit--retrieve--revise workflow.}
    Left: WSI thumbnail and patch clustering, with PRISM providing the initial global context.
    Middle/Right: across QC rounds, QCAgent issues missing-field queries, retrieves supporting patches, and revises the report, improving completeness while marking non-image-verifiable items as \emph{undetermined}.}
    \label{fig:iterative_qc_case}
\end{figure}

\noindent \textbf{Qualitative comparison.} QCAgent improves clinical completeness. Compared with PRISM draft, QCAgent adds morphology-grounded descriptions for previously missing fields and avoids unsupported claims by marking non-image-verifiable information as undetermined (Fig.~\ref{fig:qualitative}); a representative step-by-step case study is shown in Fig.~\ref{fig:iterative_qc_case}.

\subsection{Ablation Study}
\label{sec:ablation}
To evaluate the contribution of key components (Clustering, Retrieval, and Patho-R1), we conducted an ablation study on 100 randomly selected WSIs from the TCGA-STAD dataset. During preliminary evaluations, we observed a pronounced ``length bias''. Specifically, the complete framework (Full) generates highly detailed reports. While clinically informative, this verbosity heavily penalizes precision-based metrics like $F_1$-score. Conversely, configurations lacking fine-grained details achieve spuriously high $F_1$-scores. To objectively measure the coverage of core pathological features, we employ recall-centric metrics for this analysis.

\begin{table}[t]
  \centering
  \caption{Ablation study results using recall-centric metrics on 100 WSIs.}
  \label{tab:ablation}
  \small
  \begin{tabular}{lrrrrrr}
    \toprule
    {Configuration} & {AvgLen} & {R1-Recall} & {R2-Recall} & {RL-Recall} & {BS-Recall} & {FieldRecall} \\
    \midrule
    {Full (Ours)} & {130} & \textbf{0.5706} & \textbf{0.3061} & \textbf{0.5139} & 0.3703 & 0.6690 \\
    w/o Retrieval & 138 & 0.5561 & 0.2652 & 0.4809 & 0.3423 & 0.6440 \\
    w/o Patho-R1 & 98 & 0.5310 & 0.2696 & 0.4495 & \textbf{0.3986} & \textbf{0.7306} \\
    \bottomrule
  \end{tabular}
\end{table}

Ablation experiments confirm that each component is important (Table~\ref{tab:ablation}).
(a) Removing Patho-R1 (visual feature extraction) module causes the most severe performance degradation  alongside a significant reduction in report length. This demonstrates the critical role of Patho-R1 in extracting fine-grained histological features.

(b) Removing the CONCH retrieval module reduces ROUGE-L Recall by 6.4\%. This validates that dynamically retrieving local WSI regions based on missing fields effectively recovers omitted information.

\section{Conclusion}

We present an agentic, quality-controllable framework for WSI-based pathology report generation. Through an audit--retrieve--revise loop with cross-modal patch retrieval and evidence-driven revision, the QCAgent improves completeness while enforcing anti-hallucination constraints that distinguish verifiable findings from administrative items. On TCGA-STAD and an in-house gastric cancer cohort, our method consistently outperforms PRISM across overlap, semantic similarity, and Field Recall in a zero-shot, training-free setting. Future work will add expert evaluation and optimize clinical deployment.

\bibliographystyle{splncs04}
\bibliography{ref}
%


\end{document}